# ClassificationAlgorithm of Speech Data of Parkinson's Disease Based on Convolution Sparse Kernel Transfer Learning with Optimal Kernel and Parallel Sample/Feature Selection

Xiaoheng Zhang, Yongming Li* Member, IEEE, Pin Wang, Xiaoheng Tan, and Yuchuan Liu

***Abstract*—Labeled speech data from patients with Parkinson's disease (PD) are scarce, and the statistical distributions of training and test data differ significantly in the existing datasets. To solve these problems, dimensional reduction and sample augmentation must be considered. In this paper, a novel PD classification algorithm based on sparse kernel transfer learning combined with a parallel optimization of samples and features is proposed. Sparse transfer learning is used to extract effective structural information of PD speech features from public datasets as source domain data, and the fast ADDM iteration is improved to enhance the information extraction performance. To implement the parallel optimization, the potential relationships between samples and features are considered to obtain high-quality combined features. First, features are extracted from a specific public speech dataset to construct a feature dataset as the source domain. Then, the PD target domain, including the training and test datasets, is encoded by convolution sparse coding, which can extract more in-depth information. Next, parallel optimization is implemented. To further improve the classification performance, a convolution kernel optimization mechanism is designed. Using two representative public datasets and one self-constructed dataset, the experiments compare over thirty relevant algorithms. The results show that when taking the Sakar dataset, MaxLittle dataset and DNSH dataset as target domains, the proposed algorithm achieves obvious improvements in classification accuracy. The study also found large improvements in the algorithms in this paper compared with nontransfer learning approaches, demonstrating that transfer learning is both more effective and has a more acceptable time cost.**

Manuscript received December XX, 2020. This work was supported in part by NSFC under Grant 61771080,in part by the Fundamental Research Funds for the Central Universities under Grants 2019CDCGTX306 and 2019CDQYTX019,in part by the Basic and Advanced Research Project in Chongqing under Grants cstc2018jcyjAX0779 and cstc2018jcyjA3022, in part by the Chongqing Social Science Planning Project under Grant 2018YBYY133.

*Corresponding author: Yongming Li, email: yongmingli@cqu.edu.cn.

Xiaoheng Zhang is with Chongqing Radio and TV University, Chongqing 400052, China, (e-mail: 7818320@ qq.com).

Y. Li, P. Wang and Y. Liu are with the School of Microelectronics and Communication Engineering, Chongqing University, Chongqing 400030, China. (e-mail: yongmingli@cqu.edu.cn; wangpin@cqu.edu.cn; liuyc@cqu.edu.cn).

## I. Introduction

PARKINSON'S disease (PD) is a degenerative neurological disease in which physical monitoring, early diagnosis and timely intervention are the key factors for improving the quality of PD diagnosis and treatment. Speech-based PD detection has several advantages, including convenience, a high cost-performance ratio, and noncontact and noninvasive administration. Therefore, further study of speech-based PD diagnostic ability has high scientific value and practical significance [1]–[3].However, the current relevant classification algorithms for Parkinson's diagnosis are unsatisfying; thus, it is necessary to improve their accuracy.

In speech-based PD diagnosis, features are extracted from clean speech as primitive materials. In the early stage, the commonly used traditional features are vowels, rhythm, intonation and syllables, which reflect speech characteristics, and these features were most often used in isolation. The prosodic features of speech in PD were studied by Pettorino *et al.*[4]; quantitative prosody analysis was carried out, and the potential correlations between monophonic pitch, loudness and phonological discontinuities in dysarthria were further analyzed[5]. The continuous vowel information in dysarthria and the syllabic characteristics of speech in PD were studied[6], [7].

As the research progressed, the main types of features extracted from PD speech were extended to include more accurate features such as pitch, MFCC, cepstral, shimmer, jitter, HNR[8]–[12], NSR and combinations of these features. Jitter, shimmer and NSR are used to calculate classification accuracy and UPDRS score, which improved the experimental results [13], [14]. More complete speech features of PD were extracted and studied by Sakar*et al.* [15],who achieved some obtained. In addition, some of the methods for detecting PD are based on mel-cepstrum and other spectrum information[16], [17]. The biomechanical characteristics of PD patient pronunciation were adopted to make predictions based on Bayesian analysis, and the correlations between biomechanical properties, chatter and pronunciation caused by disarticulation were investigated, providing clues for the biomarkers of Parkinson's disease[18].

The methods used for feature selection include PCA[19]–[21], LDA[22], [23], NN[24]–[27], a serial search of a rough set [28] and evolutionary computation[23], [29]–[32]. The classifiers mainly include SVM[9], [10], [17], [33], KNN[21], [34], RF[9], [35]–[37], ensemble learning[38]–[40] and decision trees[41].The deep learning methods include DBN[42], DNN[43], autoencoders [44], and so on. Fuzzy theory[21], [32], [45], [46] has also been adopted as an auxiliary method.

Currently, most relevant research methods are based on two UCI datasets. One is supported by Little *et al.*[8], and the other is supported by Sakar *et al.*[15].The latter is more challenging since it originated later and accuracy on that dataset is currently lower. However, considering the regional differences among PD patients and the number of Chinese PD patients, it would be better if Chinese PD patients were considered.

It is worth noting that all the above studies were based on the current speech data and a machine learning algorithm was adopted to achieve PD classification[47], [48]. Because these methods are all directly based on a single small speech dataset, small sample-size problems prevent the algorithms from improving their performances. The best solution would be to increase the sample size while simultaneously reducing the feature dimensions. Therefore, a concrete algorithm is proposed in this paper. Convolution sparse coding(CSC)[49]–[51], which has become popular in two-dimensional image processing tasks in recent years, has good unsupervised sparse learning ability and is used to effectively compress feature dimensions to solve the small sample problem. Moreover, considering the large public speech datasets, transfer learning can be combined with sparse coding[52] to extract more valuable information from public datasets, thereby further solving both the small sample problem and the problem of different sample distributions between different datasets. Because the public speech datasets do not include labels, the CSC can be used to learn convolution kernels from these public datasets to transform them into feature map information. In this study, a parallel sample/feature selection mechanism is designed and introduced into the subsequent sparse coding processing to simultaneously obtain high-quality samples and features. In addition, to further improve the classification accuracy, a convolution kernel optimization mechanism is adopted to select the proper kernel.

The main contributions and innovations of this paper are as follows:

1) Convolution sparse coding and transfer learning are combined to solve the small sample problem of PD speech data.
2) State of the art sparse methods are optimized and introduced into PD speech data classification.
3) A parallel sample/feature selection mechanism is designed and implemented after the convolution sparse coding and transfer learning to further improve the quality of samples and features.
4) A convolution kernel optimization mechanism is proposed to enhance the current CSC algorithm.
5) Three representative PD speech datasets are considered for this study and used to verify the proposed classification algorithm. The Sakar and MaxLittle datasets are adopted as the well-known PD speech datasets (foreign PD patients) and the DNSH dataset (self=designed PD speech dataset of Chinese PD patients) is also adopted. The Sakar dataset and MaxLittle datasets involve classifications of both normal and PD patients, while the DNSH dataset involves classifications of PD patients before and after treatment. The Sakar and MaxLittle datasets are publicly available UCI datasets, and the DNSH dataset was constructed by the authors.

The remainder of this paper is divided into four main sections. Section 1 describes the background, motivation and value of this paper. Section 2 elaborates on the proposed method. Section 3 analyzes the experimental results. Section 4 discusses the contributions and limitations of this study and suggests future work.

## II. Method - PD Classification Algorithm based on Convolution Sparse Kernel Transfer Learning (PD_CSTLOK&S2)

### A. Brief Description of the Proposed Algorithm

The main flow of this algorithm is as follows. First, the public speech dataset is expanded using noise injection, forming a larger dataset. Second, the features are extracted from the data, thereby constructing a speech feature dataset as the source domain. Then, CSC learning is carried out for the source domain datasets, and a kernel matrix $\hat{\boldsymbol{D}}$ is obtained. Based on the kernels, the target domain dataset is encoded to calculate the feature maps; then, they are normalized to form a norm feature map matrix E. To improve the computational complexity and the engineering efficiency, the CSC[53] is optimized. Then, the sparse representation feature matrix is extended. Using the Relief algorithm, the weights of the mixed features in the training sample set are calculated, and feature optimization is conducted to generate a new dataset, P[48]. For descriptive simplicity, the transfer learning with an optimal kernel, simple transfer learning and nontransfer learning are denoted as PD_CSTLOK&S2 (PD classification algorithm based on convolution sparse transfer learning with optimal kernel and parallel selection), PD_CSTL&S2 (PD classification algorithm based on convolution sparse transfer learning and parallel selection) and PD_CSC&S2 (PD classification algorithm based on convolution sparse coding and parallel selection), respectively.

### B. Notation

Target domain dataset $\boldsymbol{A}=\begin{bmatrix}\vec{\boldsymbol{A}}_1\\ \vec{\boldsymbol{A}}_2\\ \vdots\\ \vec{\boldsymbol{A}}_H\end{bmatrix}=\begin{bmatrix}a_{11} & a_{12} & \cdots & a_{1N}\\ a_{21} & a_{22} & \cdots & a_{2N}\\ \cdots & \cdots & \cdots & \cdots\\ a_{H1} & a_{H2} & \cdots & a_{HN}\end{bmatrix}=\begin{bmatrix}\widetilde{\boldsymbol{A}}_1 \\ \widetilde{\boldsymbol{A}}_2\\ \vdots\\ \widetilde{\boldsymbol{A}}_M\end{bmatrix}$,

where $\vec{A}_i = [a_{i1}, a_{i2}, \cdots, a_{iN}], 1 \le i \le H$ ,label vector $C = \begin{bmatrix} c_1 \\ c_2 \\ \vdots \\ c_H \end{bmatrix}$ ,

Partitioned matrix $\widetilde{A}_i = \begin{bmatrix} a_{11} & a_{12} & \cdots & a_{1N} \\ a_{21} & a_{22} & \cdots & a_{2N} \\ \cdots & \cdots & \cdots & \cdots \\ a_{H_0 1} & a_{H_0 2} & \cdots & a_{H_0 N} \end{bmatrix}, 1 \le i \le M$ . The total number of samples is H, and the number of features per sample(Number of vector components) is N, All the samples belong to subject M, that is, the number of samples for each subject is H0=H/M; the public dataset is extended to a larger scale by injecting different SNRs and different types of noise.

The extended datasets are expressed as follows:

$$S' = \begin{bmatrix} \vec{S}'_1 \\ \vec{S}'_2 \\ \vdots \\ \vec{S}'_J \end{bmatrix} = \begin{bmatrix} \varphi(\vec{S}, \vec{N}_1, SNR_1) \\ \varphi(\vec{S}, \vec{N}_2, SNR_2) \\ \vdots \\ \varphi(\vec{S}, \vec{N}_J, SNR_J) \end{bmatrix}, \tag{1}$$

where $\vec{S}$ is the original voice signal from the public data set, $\vec{N}_j$ represent different types of noise signals, and $\varphi(\bullet)$ is a function of the type of noise adjustment and the signal-to-noise ratio(SNR).

Features are extracted for the extended data sets to form a new feature dataset

$$Y = \begin{bmatrix} \vec{Y}_1 \\ \vec{Y}_2 \\ \vdots \\ \vec{Y}_L \end{bmatrix} = \begin{bmatrix} \xi_1(\vec{S}'_1) & \xi_2(\vec{S}'_1) & \cdots & \xi_N(\vec{S}'_1) \\ \xi_1(\vec{S}'_2) & \xi_2(\vec{S}'_2) & \cdots & \xi_N(\vec{S}'_2) \\ \vdots & \vdots & \cdots & \vdots \\ \xi_1(\vec{S}'_L) & \xi_2(\vec{S}'_L) & \cdots & \xi_N(\vec{S}'_L) \end{bmatrix} = \begin{bmatrix} \widetilde{Y}_1 \\ \widetilde{Y}_2 \\ \vdots \\ \widetilde{Y}_{L'} \end{bmatrix}. \tag{2}$$

As the source domain dataset, where $\vec{Y}_i = [\xi_1(\vec{S}'_i), \xi_2(\vec{S}'_i), \cdots, \xi_N(\vec{S}'_i)], 1 \le i \le L$ , the feature extraction method from [15] was adopted to extract N different signal features. The total number of feature samples is L, $\widetilde{Y}_i$ is a two-dimensional $H_0 \times N$ block matrix, $\vec{Y}_i$ is a sparse dictionary learning training sample, and $\widetilde{Y}_i$ represents the convolution-kernel sparse learning training samples. For comparison, in the nontransfer learning version, the training set of the target domain dataset is used directly as the object of convolution sparse learning, and the remaining algorithmic processing is similar to the corresponding transfer learning version.

### *C. Convolution Sparse Coding Learning*

In CSC, given $M$ training samples $\{x_m\}_{m=1}^{M}$ , the convolution kernel group is learned by minimizing the objective function $\{d_k\}_{i=1}^{K}$ as follows:

$$\underset{e,d}{\operatorname{argmin}} \frac{1}{2} \sum_{m=1}^{M} \left\| x_m - \sum_{k=1}^{K} d_k * e_{m,k} \right\|_2^2 + \eta \sum_{m=1}^{M} \sum_{k=1}^{K} \|e_{m,k}\|_1, \text{s.t} \|d_k\|_2^2 \le 1, \forall k = \{1, \cdots, K\}, \tag{3}$$

where $x_m = \widetilde{Y}_m$ isthe $H_0 \times N$ block matrix, $e_{m,k}$ is the $H_0 \times N$ feature map matrix, and $x_m$ is approximated by convolving it with the corresponding convolution kernel $d_k$ .The notation $*$ denotes a two-dimensional convolution and $\eta$ is a regularization factor greater than zero. The solution to the above optimization problem is based on the fundamental classical framework alternating direction method of multipliers (ADMM) [54].

Formula (3) can be re-expressed as

$$\underset{e}{\arg\min} \frac{1}{2} \|De - x\|_2^2 + \eta \|e\|_1, \text{s.t.} \|d_k\|_2^2 \le 1, \tag{4}$$

where $\sum_{k=1}^{K} d_k * e_{m,k} = De$ , $D = [D_1 \ D_2 \ \cdots D_K]$ is the corresponding vectorizable convolution operator of $[d_1 \ d_2 \ \cdots d_K]$ , and $e = [e_1^T \ e_2^T \ \cdots e_K^T]^T$ is the feature map vector.

The solution can be divided into two processes.

First, the convolution kernel is fixed to obtain the feature maps. Formula (4) can be expressed as follows.

$$\underset{e,b}{\arg\min} \frac{1}{2} \|De - x\|_2^2 + \eta \|b\|_1, \ \text{s.t.} \ e - b = 0 \tag{5}$$

In (5), $\theta_1(e) = \frac{1}{2}\|De - x\|_2^2$, $\theta_2(b) = \eta\|b\|_1$, which can be solved via ADMM iteration:

$$\begin{aligned} e^{(j+1)} &= \arg\min_e \left\{ \theta_1(e) + \frac{\rho}{2} \left\| e - b^j + u^j \right\|_2^2 \right\} = \arg\min_e \frac{1}{2} \|De - x\|_2^2 + \frac{\rho}{2} \left\| e - b^j + u^j \right\|_2^2 \\ b^{(j+1)} &= \arg\min_b \left\{ \theta_2(b) + \frac{\rho}{2} \left\| e^{(j+1)} - b + u^j \right\|_2^2 \right\} = \arg\min_b \eta \|b\|_1 + \frac{\rho}{2} \left\| e^{(j+1)} - b + u^j \right\|_2^2 \\ u^{(j+1)} &= u^j + e^{(j+1)} - b^{(j+1)} \end{aligned} \tag{6}$$

Second, the feature map is fixed to solve the convolution kernel.Formula (4) can be expressed as follows:

$$\underset{d,c}{\arg\min} \frac{1}{2} \|Ed - x\|_2^2, \ \text{s.t.} \|c_k\|_2^2 \le 1 \ \text{and} \ d - c = 0. \tag{7}$$

In (7), $\theta_1(d) = \frac{1}{2}\|Ed - x\|_2^2$ , $\theta_2(c)$ is the indicator function of the convex set $\|c_k\|_2^2 \le 1$ , and in (8), $\operatorname{prox}(\bullet)$ computes the proximal operator, which can be solved via ADMM iteration:

$$\begin{aligned} d^{(j+1)} &= \arg\min_d \left\{ \theta_1(d) + \frac{\rho}{2} \left\| d - c^j + v^j \right\|_2^2 \right\} = \arg\min_d \frac{1}{2} \|Ed - x\|_2^2 + \frac{\rho}{2} \left\| d - c^j + v^j \right\|_2^2 \\ c^{(j+1)} &= \operatorname{prox}_{\theta_2(c)}(d^{(j+1)} + v^j) \\ v^{(j+1)} &= v^j + d^{(j+1)} - c^{(j+1)} \end{aligned} \tag{8}$$

Finally, a set of sparse convolution kernels $[d'_1, d'_2, \cdots, d'_k]$ is obtained by the alternating iterations.

### *D. Solution of Fast Convolution Sparse Coding*

CSC feature map learning can be realized by ADMM. The fast convolution sparse coding algorithm [53] is adopted here when the convolution kernels $[d'_1, d'_2, \cdots, d'_k]$ are fixed. The feature extraction formula is expanded as follows:

$$\begin{aligned} e^{(j+1)} &= \arg_e \min \frac{\lambda}{2} \|De - x\|_2^2 + \frac{1}{2} \left\| e - b^j - u^j \right\|_2^2 \\ &= (\lambda D^T D + I)^{-1} (\lambda D^T x + b^j - u^j) \\ &= (I - \lambda D^T (\lambda D^T D + I)^{-1} D)(\lambda D^T x + b^j - u^j) \\ &= (I - \lambda D^T (\lambda \sum D_k D_k^T + I)^{-1} D)(\lambda D^T x + b^j - u^j) \end{aligned}, \tag{9}$$

$$\begin{aligned}\boldsymbol{b}^{(j+1)} &= \arg_{b}\min \alpha \|\boldsymbol{b}\|_1 + \frac{1}{2}\left\|\boldsymbol{e}^{(j+1)} - \boldsymbol{b} + \boldsymbol{u}^{(j)}\right\|_2^2 \\ &= \boldsymbol{S}_{\alpha}\left(\left\|\boldsymbol{e}^{(j+1)} + \boldsymbol{u}^{(j)}\right\|\right) \\ &= \begin{cases} \boldsymbol{e}^{(j+1)} + \boldsymbol{u}^{(j)} - \alpha, \boldsymbol{e}^{(j+1)} + \boldsymbol{u}^{(j)} > \alpha \\ 0, \left|\boldsymbol{e}^{(j+1)} + \boldsymbol{u}^{(j)}\right| \le \alpha \\ \boldsymbol{e}^{(j+1)} + \boldsymbol{u}^{(j)} + \alpha, \ \boldsymbol{e}^{(j+1)} + \boldsymbol{u}^{(j)} < \alpha \end{cases},\end{aligned} \quad (10)$$

$$\boldsymbol{u}^{(j+1)} = \boldsymbol{u}^{j} + \boldsymbol{e}^{(j+1)} - \boldsymbol{b}^{(j+1)}. \quad (11)$$

Because matrix inversion is extremely time consuming, in (9), $\left(I + \lambda \sum \boldsymbol{D}_k \boldsymbol{D}_k^T\right)^{-1}$ can be precomputed in the Fourier domain by $\frac{1}{1 + \sum_{k=1}^{K} \left|\hat{d}_k\right|^2}$ for the fixed parameters to greatly reduce the algorithm complexity, where $\hat{d}_k$ is the magnitudes of the Fourier transform of $d_k$. In (10), $S_{\alpha}(\bullet)$ is the soft threshold function of the element operation. A stable feature map can be obtained after many iterations.

Similarly, the kernel learning formula is expanded as follows:

$$\begin{aligned}\boldsymbol{d}^{(j+1)} &= \arg\min_{\boldsymbol{d}} \frac{\lambda}{2}\|\boldsymbol{E}\boldsymbol{d} - \boldsymbol{x}\|_2^2 + \frac{1}{2}\left\|\boldsymbol{d} - \boldsymbol{c}^{j} + \boldsymbol{v}^{j}\right\|_2^2 \\ &= \left(I - \lambda \boldsymbol{E}^T\left(\lambda \boldsymbol{E}\boldsymbol{E}^T + I\right)^{-1}\boldsymbol{E}\right)\left(\lambda \boldsymbol{E}^T\boldsymbol{x} + \boldsymbol{c}^{j} - \boldsymbol{v}^{j}\right) \\ &= \left(I - \lambda \boldsymbol{E}^T\left(\lambda \sum \boldsymbol{E}_k \boldsymbol{E}_k^T + I\right)^{-1}\boldsymbol{E}\right)\left(\lambda \boldsymbol{E}^T\boldsymbol{x} + \boldsymbol{c}^{j} - \boldsymbol{v}^{j}\right)\end{aligned}, \quad (12)$$

$$\boldsymbol{c}^{(j+1)} = \begin{cases} (\boldsymbol{d}^{j+1} + \boldsymbol{v}^{j})\mathrm{F}_{\mathrm{supp}}(\boldsymbol{d}^{j+1}) / \left\|\boldsymbol{d}^{j+1} + \boldsymbol{v}^{j}\right\|_2^2, & \text{if } \left\|\boldsymbol{d}^{j+1} + \boldsymbol{v}^{j}\right\|\mathrm{F}_{\mathrm{supp}}(\boldsymbol{d}^{j+1}) > 1 \\ (\boldsymbol{d}^{j+1} + \boldsymbol{v}^{j})\mathrm{F}_{\mathrm{supp}}(\boldsymbol{d}^{j+1}), & \text{otherwise} \end{cases}, \quad (13)$$

where $\mathrm{F}_{\mathrm{supp}}(\boldsymbol{d})$ is a mask that takes a 1 on $\mathrm{SUPP}(\boldsymbol{d})$ and 0 otherwise:

$$\boldsymbol{v}^{(j+1)} = \boldsymbol{v}^{j} + \boldsymbol{d}^{(j+1)} - \boldsymbol{c}^{(j+1)}. \quad (14)$$

To further improve the efficiency of the CSC algorithm, by revisiting the iterative scheme (9)-(14), it can be observed that the variables $e$ and $d$ play only an intermediate role. Based on [55], a new iterative feature extraction formula is generated as follows:

$$\text{step1.} \begin{cases} \bar{\boldsymbol{e}}^{j} = \left(I - \lambda \boldsymbol{D}^T\left(I + \lambda \sum \boldsymbol{D}_k \boldsymbol{D}_k^T\right)^{-1}\boldsymbol{D}\right)\left(\lambda \boldsymbol{D}^T\boldsymbol{x} + \boldsymbol{b}^{j} - \boldsymbol{u}^{j}\right) \\ \bar{\boldsymbol{u}}^{j} = \boldsymbol{u}^{j} + \bar{\boldsymbol{e}}^{j} - \boldsymbol{b}^{j} \\ \bar{\boldsymbol{b}}^{j} = \begin{cases} \bar{\boldsymbol{e}}^{j} + \bar{\boldsymbol{u}}^{(j)} - \alpha & \bar{\boldsymbol{e}}^{j} + \bar{\boldsymbol{u}}^{(j)} > \alpha \\ 0 & \left|\bar{\boldsymbol{e}}^{j} + \bar{\boldsymbol{u}}^{(j)}\right| \le \alpha \\ \bar{\boldsymbol{e}}^{j} + \bar{\boldsymbol{u}}^{(j)} + \alpha & \bar{\boldsymbol{e}}^{j} + \bar{\boldsymbol{u}}^{(j)} < \alpha \end{cases} \end{cases} . \quad (15)$$

$$\text{step2.} \begin{pmatrix} \boldsymbol{b}^{(j+1)} \\ \boldsymbol{u}^{(j+1)} \end{pmatrix} = \begin{pmatrix} \boldsymbol{b}^{j} \\ \boldsymbol{u}^{j} \end{pmatrix} - \gamma \begin{pmatrix} \boldsymbol{b}^{j} - \bar{\boldsymbol{b}}^{j} \\ \boldsymbol{u}^{j} - \bar{\boldsymbol{u}}^{j} \end{pmatrix}$$

Then, a new kernel learning formula is generated as follows:

$$\text{step1.} \begin{cases} \bar{\boldsymbol{d}}^{j} = \left(I - \lambda \boldsymbol{E}^T\left(I + \lambda \sum \boldsymbol{E}_k \boldsymbol{E}_k^T\right)^{-1}\boldsymbol{E}\right)\left(\lambda \boldsymbol{E}^T\boldsymbol{x} + \boldsymbol{c}^{j} - \boldsymbol{v}^{j}\right) \\ \bar{\boldsymbol{v}}^{j} = \boldsymbol{v}^{j} + \bar{\boldsymbol{d}}^{j} - \boldsymbol{c}^{j} \\ \bar{\boldsymbol{c}}^{j} = \begin{cases} (\bar{\boldsymbol{d}}^{j} + \bar{\boldsymbol{v}}^{j})\mathrm{F}_{\mathrm{supp}}(\bar{\boldsymbol{d}}^{j}) / \left\|\bar{\boldsymbol{d}}^{j} + \bar{\boldsymbol{v}}^{j}\right\| & \text{if } \left\|\bar{\boldsymbol{d}}^{j} + \bar{\boldsymbol{v}}^{j}\right\|\mathrm{F}_{\mathrm{supp}}(\hat{\boldsymbol{d}}^{j}) > 1 \\ (\bar{\boldsymbol{d}}^{j} + \bar{\boldsymbol{v}}^{j})\mathrm{F}_{\mathrm{supp}}(\bar{\boldsymbol{d}}^{j}) & \text{otherwise} \end{cases}, \end{cases}$$

$$\text{step2.} \begin{pmatrix} \boldsymbol{c}^{(j+1)} \\ \boldsymbol{v}^{(j+1)} \end{pmatrix} = \begin{pmatrix} \boldsymbol{c}^{j} \\ \boldsymbol{v}^{j} \end{pmatrix} - \gamma \begin{pmatrix} \boldsymbol{c}^{j} - \bar{\boldsymbol{c}}^{j} \\ \boldsymbol{v}^{j} - \bar{\boldsymbol{v}}^{j} \end{pmatrix} \quad (16)$$

where $\gamma$ is a relaxation factor.

A pseudocode description of the fast kernel learning algorithm from a public dataset is shown in Algorithm 1.

***Algorithm 1: Fast kernel learning algorithm***

**Input:** Public dataset $\vec{S}$, total sample size $H$, number of features per sample $N$

**Output:** convolution kernel $\hat{D}$

**Procedure:**

1.Using Formula (1), add different types of noise with different signal-to-noise ratios to the dataset and extend it to the set $\boldsymbol{S}$;

2.Using Formula (2), extract the features of the speech samples from $\boldsymbol{S}$ and construct the feature dataset, which is the source domain dataset $\boldsymbol{Y}$;

3. **for** iteration = 1…iter_num **do**

4. **for** iteration1 = 1…iter_num1 **do**

5.Using Formula (15), fix the convolution kernel to obtain the feature maps.

6. **end for**

7.**for** iteration2 = 1…iter_num2 **do**

8.Using Formula (16), fix the feature map to obtain the convolution kernel.

9.**end for**

10:**end for**

*E. Algorithm for Simultaneous Selection of Speech Samples and Features*

In Formula (16), by replacing $\boldsymbol{x}$ with $\tilde{A}_i$ and $\boldsymbol{D}$ with $\hat{\boldsymbol{D}}$, the target domain feature matrix $\tilde{A}_i$ is mapped to the feature map matrix $\boldsymbol{e} = \lfloor e_1 \ e_2 \ \cdots e_K \rfloor$ through a finite number of iterations. Then, $e_k$ is selected as a specific mapping $\tilde{\boldsymbol{E}}_i$, and a new feature set $\boldsymbol{E} = \begin{bmatrix} \tilde{\boldsymbol{E}}_1 \\ \tilde{\boldsymbol{E}}_2 \\ \vdots \\ \tilde{\boldsymbol{E}}_M \end{bmatrix}$ is constructed.

Fig. 1 shows the images in the processing flow of CSC for the DNSH dataset. In Fig. 1(a), the locally normalized image is a visualization of one subject's data from the DNSH dataset. The image in Fig.1(b)is the sparse coding of the subject data in Fig. 1(a), obtained from the convolutional sparse model. Fig. 1(c) shows a feature map extracted from the subject data in Fig. 1(a). The sparsity of image Fig.1(b) is greater than that of image Fig.1(a), and Fig.1(c) is extremely sparse, which means that the feature extracted from convolutional sparse coding are highly significant.

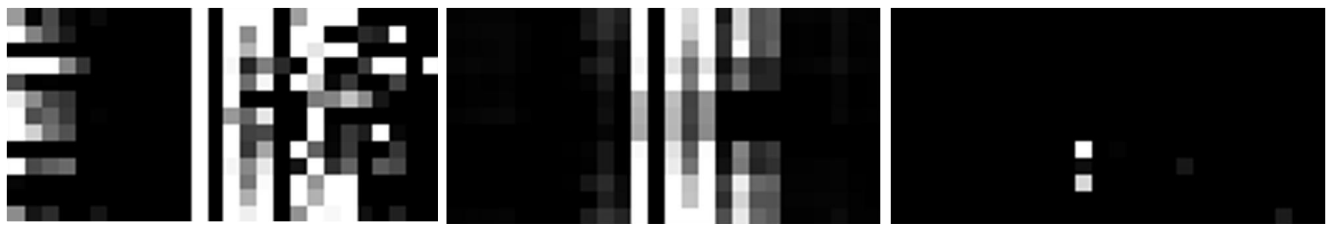

Fig.1.Images in the processing flow of CSC for the DNSH dataset: (a) Locally normalized image.;(b)Image obtained from the convolution sparse model; (c) Feature map.

The feature map matrix $\boldsymbol{E}$ is extended to $\boldsymbol{G}$ as follows:

$$\boldsymbol{G}=\begin{bmatrix}\vec{\boldsymbol{G}}_1\\ \vec{\boldsymbol{G}}_2\\ \vdots\\ \vec{\boldsymbol{G}}_M\end{bmatrix}=\begin{bmatrix}\textbf{RESHAPE}\left(\widetilde{\boldsymbol{E}}_1 \quad H_0\times N \quad 1\times N'\right)\\ \textbf{RESHAPE}\left(\widetilde{\boldsymbol{E}}_2 \quad H_0\times N \quad 1\times N'\right)\\ \vdots\\ \textbf{RESHAPE}\left(\widetilde{\boldsymbol{E}}_M \quad H_0\times N \quad 1\times N'\right)\end{bmatrix}=\begin{bmatrix}\gamma_{11} & \gamma_{12} & \cdots & \gamma_{1N_0}\\ \gamma_{21} & \gamma_{22} & \cdots & \gamma_{2N_0}\\ \cdots & \cdots & \cdots & \cdots\\ \gamma_{M1} & \gamma_{M2} & \cdots & \gamma_{MN_0}\end{bmatrix} \quad (17$$

where the feature extension of CSC expands $H_0$ row vectors $\vec{\boldsymbol{E}}_i$ of the same subject into one row vector; the convolution sparse coding feature expansion aims to transform the feature matrix $\widetilde{\boldsymbol{E}}_i$ into one row vector, normalize $\boldsymbol{G}$ to obtain $\boldsymbol{G}'$, and split the matrix $\boldsymbol{G}'=\begin{bmatrix}\boldsymbol{G_m}\\ \boldsymbol{T_m}\end{bmatrix}$ into training sets $\boldsymbol{G_m}==\left[\vec{\boldsymbol{\Gamma}}_{G1} \quad \vec{\boldsymbol{\Gamma}}_{G2} \quad \cdots \quad \vec{\boldsymbol{\Gamma}}_{GN_0}\right]$ and test sets $\boldsymbol{T_m}$.

Based on the Relief algorithm, the weight $\vec{\boldsymbol{W}}=\left[w_1 \ w_2 \ \cdots w_{N_0}\right]$ of the normalized feature vector array $\left[\vec{\boldsymbol{\Gamma}}'_{G1} \quad \vec{\boldsymbol{\Gamma}}'_{G2} \quad \cdots \quad \vec{\boldsymbol{\Gamma}}'_{GN_0}\right]$ is calculated. The weight of feature $\vec{\boldsymbol{\Gamma}}'_j$ is as follows:

$$w_j=\sum_i^M\left(-\sum_r^R\left\|\gamma_{ij}-\mathrm{M}_r(\gamma_{ij})\right\|^2+\sum_r^R\left\|\gamma_{ij}-\mathrm{H}_r(\gamma_{ij})\right\|^2\right), \quad (18)$$

where $\mathrm{M}(\gamma_{ij})$ is the neighbor set of the same class of $\gamma_{ij}$, $R$ is the set number, $H(\gamma_{ij})$ is the neighbor set of the other class of $\gamma_{ij}$, and $M_r(\gamma_{ij})\in M(\gamma_{ij})$, $H_r(\gamma_{ij})\in H(\gamma_{ij})$ .By reordering $\vec{\boldsymbol{W}}$ and so that $w_1\geq w_2\geq\cdots\geq w_Q$, the feature sets $\boldsymbol{P}^{Q}_{\boldsymbol{G_m}}$ and $\boldsymbol{P}^{Q}_{\boldsymbol{T_m}}$ are reconstructed according to the weights.

$$\boldsymbol{P}=\begin{bmatrix}\boldsymbol{P}^{Q}_{\boldsymbol{G_m}}\\ \boldsymbol{P}^{Q}_{\boldsymbol{T_m}}\end{bmatrix}=\begin{bmatrix}\vec{\boldsymbol{P}}_1\\ \vec{\boldsymbol{P}}_2\\ \vdots\\ \vec{\boldsymbol{P}}_M\end{bmatrix}=\left[\begin{bmatrix}\vec{\boldsymbol{\Gamma}}''_{G_1}\\ \vec{\boldsymbol{\Gamma}}''_{T_1}\end{bmatrix}\begin{bmatrix}\vec{\boldsymbol{\Gamma}}''_{G_2}\\ \vec{\boldsymbol{\Gamma}}''_{T_2}\end{bmatrix}\cdots\begin{bmatrix}\vec{\boldsymbol{\Gamma}}''_{G_Q}\\ \vec{\boldsymbol{\Gamma}}''_{T_Q}\end{bmatrix}\right]. \quad (19)$$

$$=\left[\begin{bmatrix}\vec{\boldsymbol{\Gamma}}'_{G_1}\\ \vec{\boldsymbol{\Gamma}}'_{T_1}\end{bmatrix}\begin{bmatrix}\vec{\boldsymbol{\Gamma}}'_{G_2}\\ \vec{\boldsymbol{\Gamma}}'_{T_2}\end{bmatrix}\cdots\begin{bmatrix}\vec{\boldsymbol{\Gamma}}'_{G_{N0}}\\ \vec{\boldsymbol{\Gamma}}'_{T_{N0}}\end{bmatrix}\right]\left[\vec{l}_1^{\,T} \ \vec{l}_2^{\,T}\cdots \ \vec{l}_Q^{\,T}\right]$$

The $Q$ dimension vectors $\left[\vec{\boldsymbol{\Gamma}}'_{G1} \quad \vec{\boldsymbol{\Gamma}}'_{G2} \quad \cdots \quad \vec{\boldsymbol{\Gamma}}'_{GQ}\right]$ and $\left[\vec{\boldsymbol{\Gamma}}'_{T1} \quad \vec{\boldsymbol{\Gamma}}'_{T2} \quad \cdots \quad \vec{\boldsymbol{\Gamma}}'_{TQ}\right]$ are selected according to the maximum $Q$ weights of $\left[\vec{\boldsymbol{\Gamma}}'_{G1} \quad \vec{\boldsymbol{\Gamma}}'_{G2} \quad \cdots \quad \vec{\boldsymbol{\Gamma}}'_{GN_0}\right]$, $\vec{l}_i=\left[\underbrace{0\cdots 0\,1}_{index}\,0\cdots 0\right]$, and $index$ is the column marker for the feature column vector with $w_i$.

The pseudocode description for the simultaneous selection based on the convolution kernel algorithm is shown in Algorithm 2.

***Algorithm 2:Simultaneous Selection based on Convolution Kernel Algorithm(SS_CK)***

**Input:** Convolution kernel $\hat{\boldsymbol{D}}$, target domain data set $\boldsymbol{A}$, total sample size $H$, number of features per sample $N$, number of subjects $M$

**Output:** accuracy

**Procedure:**

1: Compute the feature map matrix $\boldsymbol{E}$ of the target domain dataset, expand $\boldsymbol{G}$, and normalize it to $\boldsymbol{G}'$;

2: Split $\boldsymbol{G}'=\begin{bmatrix}\boldsymbol{G_m}\\ \boldsymbol{T_m}\end{bmatrix}$ into a training set $\boldsymbol{G_m}$ and a test set $\boldsymbol{T_m}$;

3: Scan the sample $\vec{\boldsymbol{P}}_{\boldsymbol{G_m}(i)}$ of training set $\boldsymbol{G_m}$, and select $R$ neighbors from the same type of samples $\vec{\boldsymbol{P}}_{\boldsymbol{G_m}(i)}$ to construct $M(\gamma_{ij})$. Then, select $R$ neighbors from the other type of samples $\vec{\boldsymbol{P}}_{\boldsymbol{G_m}(i)}$ to construct $\mathrm{H}(\gamma_{ij})$;

4: Using (22), calculate $\vec{\boldsymbol{W}}$;

5: According to the maximized $Q$ weights in $\vec{\boldsymbol{W}}$ ,select the optimal features and obtain a new matrix $\boldsymbol{P}$;

6: Perform LOSO CV; then, the classification accuracy is calculated based on an SVM classifier with a linear kernel.

## *F. Convolution Kernel Optimization*

The classification quality is influenced by numerous factors, such as the number of convolution kernels, the number of feature maps, and the number of features selected by simultaneous selection; therefore, it is important to train a proper sparse coding kernel. In (20), both the sparse convolution characteristics of Algorithm 1 and the classification accuracy of Algorithm 2 are considered. $\min\limits_{a,d}\frac{1}{2}\sum_{m=1}^{M}\left\|\boldsymbol{x}_m-\sum_{k=1}^{K}\boldsymbol{d}_k*a_{m,k}\right\|_2^2+\lambda\sum_{m=1}^{M}\sum_{k=1}^{K}\left\|a_{m,k}\right\|_1$ is a biconvex problem; consequently, it is difficult to prove whether $\textbf{CKSS}(\boldsymbol{d}_k,Q)$ is convex. Clearly, b in (20) is a complex combinatorial optimization problem with a large parameter search space:

$$\boldsymbol{d}_k=\arg\begin{bmatrix}\mathrm{a:}\quad \min\limits_{a,d}\frac{1}{2}\sum_{m=1}^{M}\left\|\boldsymbol{x}_m-\sum_{k=1}^{K}\boldsymbol{d}_k*a_{m,k}\right\|_2^2+\lambda\sum_{m=1}^{M}\sum_{k=1}^{K}\left\|a_{m,k}\right\|_1\\ \mathrm{b:}\quad \max\ \textbf{SS_CK}\left(\boldsymbol{d}_k,Q\right)\ \textbf{on kernel optimal set}\end{bmatrix}. \quad (20)$$

For the reasons above, a simplified kernel optimization solution mechanism is proposed in this paper. The core idea is that by initializing the convolution kernel $\boldsymbol{D}_{ij}$ using multiple random seeds in parallel, the optimal $\boldsymbol{D}_{\text{optimal}}$ is selected based on the classification accuracy of the kernel training set.

The pseudocode of the entire proposed algorithm, including convolution kernel optimization, is shown in Algorithm 3.

***Algorithm 3(Complete Algorithm): PD_CSTLOK&S2***

**Input:** Public dataset $\vec{S}$, target domain data set $A$, total sample size $H$, number of features per sample $N$, number of subjects $M$

**Output:** accuracy, sensitivity, specificity

**Procedure:**

Split $A$ into a kernel optimization set $A_1$ and a data training-test set $A_2$;

**for** kernel_num= 1…$i$ **do**

  **for** featuremap_num= 1…kernel_num **do**

    **parallel for** seed= 1…$j$**do**

      Randomly initialize the convolution kernel $D_{ij}$;

      Calculate the convolution kernel on dataset $A_1$, using Algorithm 1;

      Calculate the accuracy using Algorithm 2;

      Obtain $D_{\text{optimal}}$ when the best result of $\mathbf{SS_CK}\left(D_{\text{optimal}}, Q\right)$ is achieved.

    **end parallel for**

  Calculate the classification accuracy on dataset $A_2$ based on $D_{\text{optimal}}$ using Algorithm 2;

  **end for**

**end for**

## III. Experimental Results and Analysis

### A. Experimental Environment

The source domain dataset of this paper is the standard speech dataset TIMIT. The training set includes data from 40 men and 40 women, each with 3 sentences. The dataset is supplemented with noise (from the NOISEX-92 noise dataset) and expanded. The expanded dataset is 10 times larger than the original dataset. Based on the expanded dataset, 26 features (Jitter,Shimmer,AC,NTH,HTN, etc.) are extracted to construct the feature set (feature matrix). In this paper, three PD datasets were selected as the target sets. The 1$^{st}$ dataset was created by Little *et al.*[8], [9]. The 2$^{nd}$ dataset was created by Sakar *et al.*[15] in 2014. The 3$^{rd}$ dataset was collected by the authors at the Department of Neurology, Southwest Hospital and contains 10 subjects in the closing period and 21 subjects in opening period. The corresponding 13 voice samples collected arelabeled '1', '2', '3', '4', '5', '6', '7', '8', '9', '10', 'a', 'o', and 'u', and the extracted feature types are the same as the features in the Sakar dataset.

The subjects in the dataset were subjected to cross-validation by LOSO. The main reason for choosing the cross-validation method is that there are insufficient samples; cross-validation maximizes the number of training samples to better reflect the algorithm's potential. Therefore, the test accuracy is closer to the results of an actual application scenario. Different from the current literature, which uses k-fold and hold-one-out cross-validation, the training set and test set represent different subjects to ensure that the classification accuracy is both realistic and consistent with the actual diagnosis.

The hardware configuration of the experiment platform was as follows: CPU (Intel i5-3230M), 4GBof memory and MATLAB R2014a. For the PD_CSTLOK&S2 algorithm, the number of random seeds was 10, and the numbers of main training iterations, feature map iterations and convolution kernel iterations were set to 100, 10 and 10, respectively.

The relaxation factor $\gamma = 1$. The number of convolution kernels ranged from 2 to 8, and the size of the convolution kernel was 8*8 for the Sakar and DNSH datasets and 4*4 for the MaxLittle dataset.

### B. Verifying the Main Parts of the Proposed Algorithm

#### 1) Verification of Parallel Sample/FeatureSelection

Fig. 2 shows a comparison of the classification accuracy of the PD_CSTLOK&S2 algorithm before and after parallel sample/feature selection for the MaxLittle dataset. The abscissa is the (convolution kernel num, feature map num), and the ordinate is the classification accuracy. As Fig. 2 shows, the classification accuracy rate after applying parallel sample/feature selection is significantly higher than that before applying the technique. The results show that parallel sample/feature selection can significantly improve the test accuracy.

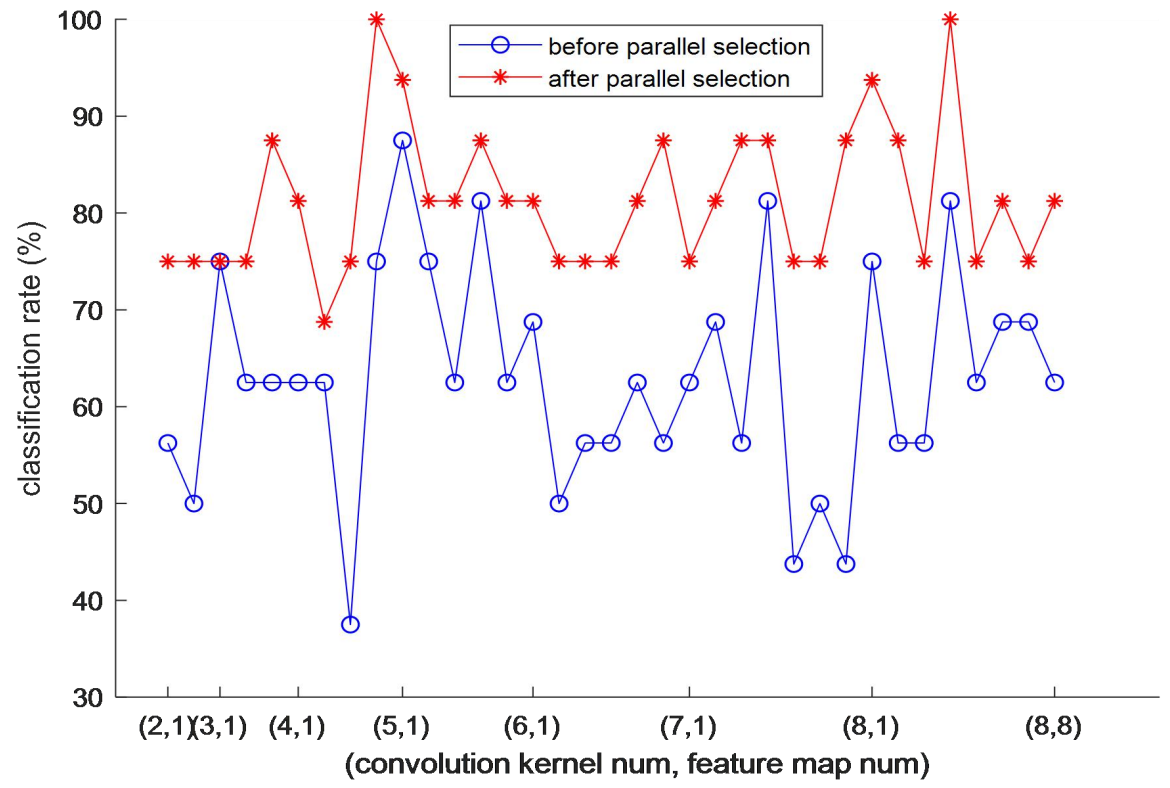

Fig.2. Comparison of PD_CSTLOK&S2 based on the algorithm before and after parallel sample/feature selection.

#### 2) Verification of Transfer Learning and Optimal Convolution Kernels

The classification performance is illustrated in Fig. 3,where the confusion matrixes show the actual and predicted classifications. Fig. 3(a) indicates the comparison results of PD_CSC&S2, PD_CSTL&S2 and PD_CSTLOK&S2 on the Sakar dataset. The case is similar to the MaxLittle dataset in Fig. 3(b) and DNSH dataset in Fig. 3(c).

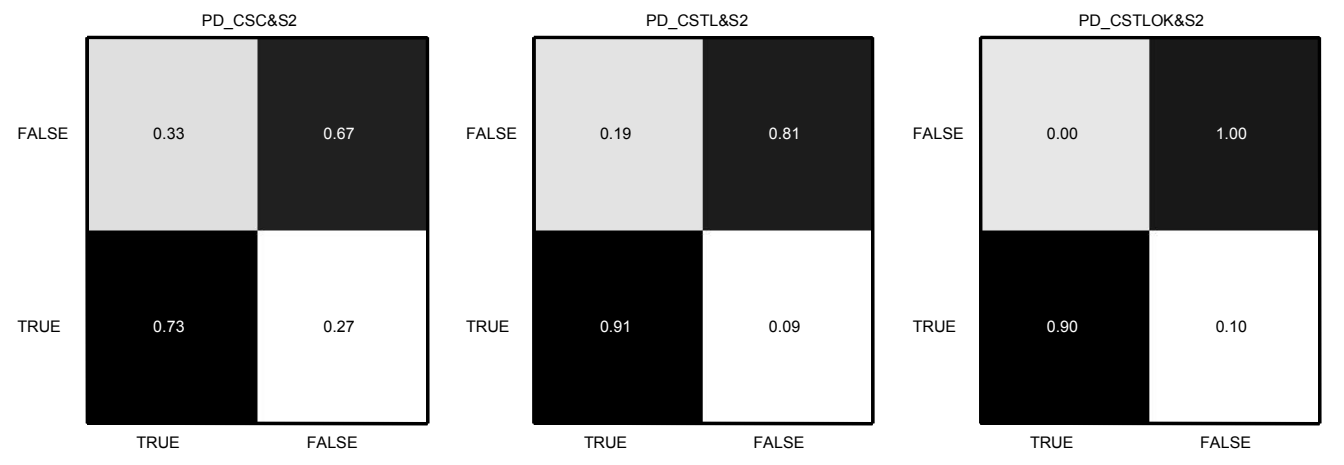

Fig.3(a). The confusion matrix for the PD_CSC&S2, PD_CSTL&S2 and PD_CSTLOK&S2 algorithms on the Sakar dataset.

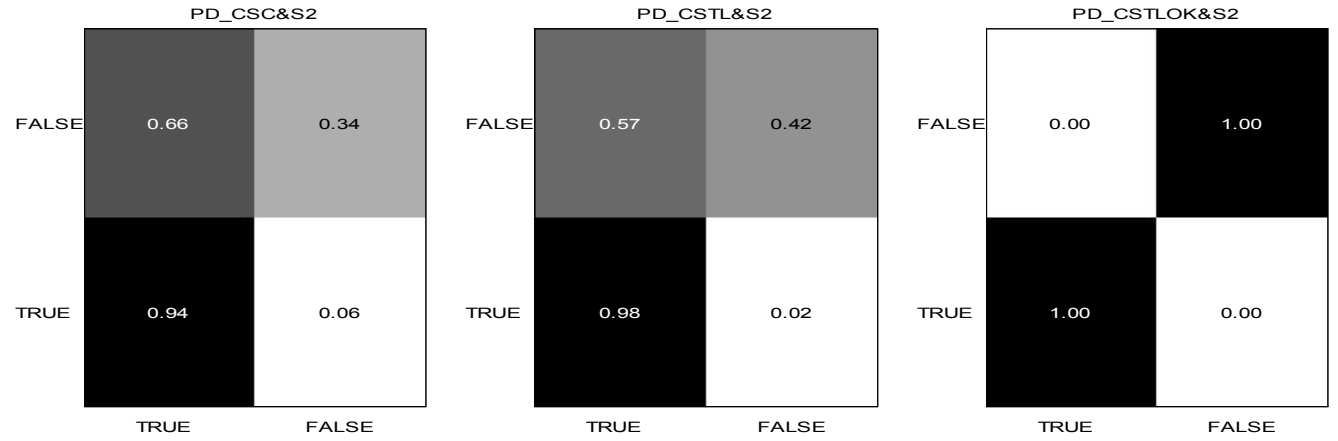

Fig.3(b). The confusion matrix for the PD_CSC&S2, PD_CSTL&S2 and PD_CSTLOK&S2 algorithms on the MaxLittle dataset.

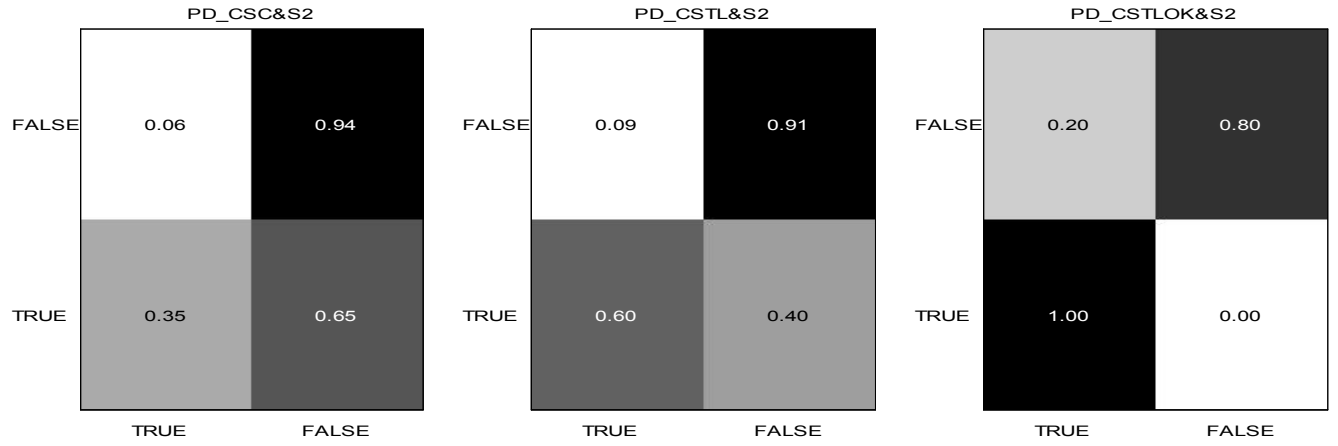

Fig.3(c). The confusion matrix for the PD_CSC&S2, PD_CSTL&S2 and PD_CSTLOK&S2 algorithms on the DNSH dataset.

As the comparison in Fig. 3(a) shows, moving from PD_CSC&S2, PD_CSTL&S2 and PD_CSTLOK&S2, the sensitivity improves from 73% to 91%, increasing by 18%, and decreases to 90% slightly. The specificity improves from 67% to 81% and then to 100%, increasing by 14% and 19%, respectively. In Fig. 3(b), the sensitivity improves from 94% to 98% and then to 100%, increasing by 4% and 2%, respectively. The specificity improves from 34% to 42% and then to 100%, increasing by 8% and 58%, respectively. In Fig. 3(c), the sensitivity improves from 35% to 60% and then to 100%, increasing by 25% and 40%, respectively.

### C. *Effects of Parameters on the Proposed Algorithm's Performance*

The convolution kernel is one of the main parameters of PD_CSTLOK&S2; therefore, it is necessary to study its effect on the algorithm's performance. For the DNSH dataset, as the number of convolution kernels increases from 2 to 8, the relationship between the number of convolution kernels and the classification accuracy is shown in Table I.

TABLE I
RELATIONSHIP BETWEEN THE NUMBER OF CONVOLUTION KERNELS AND CLASSIFICATION ACCURACY FOR THE SAKAR DATASET

| Convolution kernel num | Feature map num | Accuracy(%) | Sensitivity(%) | Specificity(%) |
|---|---|---|---|---|
| 2 | 1 | 75.0 | 80.0 | 70.0 |
| | 2 | 60.0 | 60.0 | 60.0 |
| 3 | 1 | 70.0 | 70.0 | 70.0 |
| | 2 | 65.0 | 60.0 | 70.0 |
| | 3 | 60.0 | 60.0 | 60.0 |
| 4 | 1 | 80.0 | 80.0 | 80.0 |
| | 2 | 60.0 | 60.0 | 60.0 |
| | 3 | 60.0 | 70.0 | 50.0 |
| | 4 | 85.0 | 100.0 | 70.0 |
| 5 | 1 | 70.0 | 80.0 | 60.0 |
| | 2 | 80.0 | 70.0 | 90.0 |
| | 3 | 80.0 | 80.0 | 80.0 |
| | 4 | 50.0 | 60.0 | 40.0 |
| | 5 | 85.0 | 80.0 | 90.0 |
| 6 | 1 | 60.0 | 70.0 | 50.0 |
| | 2 | 65.0 | 50.0 | 80.0 |
| | 3 | 65.0 | 60.0 | 70.0 |
| | 4 | 80.0 | 80.0 | 80.0 |
| | 5 | 60.0 | 60.0 | 60.0 |
| | 6 | 85.0 | 80.0 | 90.0 |
| 7 | 1 | 65.0 | 80.0 | 50.0 |
| | 2 | 75.0 | 80.0 | 70.0 |
| | 3 | 75.0 | 60.0 | 90.0 |
| | 4 | 80.0 | 90.0 | 70.0 |
| | 5 | 65.0 | 70.0 | 60.0 |
| | 6 | 75.0 | 80.0 | 70.0 |
| | 7 | 85.0 | 80.0 | 90.0 |
| 8 | 1 | 80.0 | 80.0 | 80.0 |
| | 2 | 80.0 | 90.0 | 70.0 |
| | 3 | 70.0 | 70.0 | 70.0 |
| | 4 | 65.0 | 70.0 | 60.0 |
| | 5 | 75.0 | 60.0 | 90.0 |
| | 6 | 90.0 | 90.0 | 90.0 |
| | 7 | 80.0 | 80.0 | 80.0 |
| | 8 | **95.0** | 90.0 | **100.0** |

As shown in Table I, with 8 convolution kernels and 8 feature maps, the average classification accuracy reaches a maximum of 95.0%, the sensitivity reaches 90.0% and the specificity reaches 100.0%. It can be found that the best result can be obtained by selecting proper convolution kernel num and feature map num.

### D. *Comparison with Representative Algorithms*

Table II presents the classification and comparison results of this algorithm on the Sakar dataset. The algorithm proposed in this paper is compared with other representative algorithms on this dataset. In addition, the proposed algorithm is compared with other relevant algorithms, including the SVM with linear and radial kernels, DBN, CNN and a deep autoencoder algorithm.

TABLE II
COMPARISON OF THE CLASSIFICATION RESULTS OF THE PROPOSED ALGORITHM (SAKAR DATASET)

| Study | Method | Accuracy(%) | Sensitivity(%) | Specificity(%) |
|---|---|---|---|---|
| Sakar *et al.*[15] | KNN+SVM | 55(LOSO CV) | 60 | 50 |
| Canturk and Karabiber [56] | 4 Feature Selection Methods+ 6 Classifiers | 57.5(LOSO CV) | 54.28 | 80 |
| Eskidere *et al.*[38] | Random Subspace Classifier Ensemble | 74.17(10-fold CV) | — | — |
| Behroozi and Sami[39] | Multiple classifier framework | 87.50(A-MCFS) | 90.00 | 85.00 |
| Zhang *et al.*[35] | MENN+RF with MENN | 81.5(LOSO CV) | 92.50 | 70.50 |
| Benba *et al.*[57] | HFCC+SVM | 87.5(LOSO CV) | 90.00 | 85.00 |
| Li *et al.*[33] | Hybrid feature learning+SVM | 82.50(LOSO CV) | 85.00 | 80.00 |
| Vadovsk and Paralic[36] | C4.5+C5.0+RF+CART | 66.5(4-fold CV) | — | — |
| Zhang [22] | LSVM+MSVM+ RSVM+CART+KNN+LDA+NB | 94.17(Hold out) | 50.00 | 94.92 |
| Benba *et al.*[17] | MFCC+SVM | 82.5(LOSO CV) | 80.0 | 85.0 |
| Kraipeerapun and Amornsamanku [58] | Stacking+CMTNN | 75(10-fold CV) | — | — |
| Khan *et al.*[40] | Evolutionary Neural Network Ensembles | 90(10-fold CV) | 93.00 | 97.00 |
| Ali *et al.*[23] | LDA-NN-GA | 95(LOSO CV) | 95 | 95 |
| - | DBN | 54.6(LOSO CV) | 52.4 | 56.8 |
| - | CNN | 60.0(LOSO CV) | 63.0 | 57.0 |
| - | DBN+SVM | 50.5(LOSO CV) | 53.0 | 48.0 |
| - | DBN+SVM(TL) | 55.5(LOSO CV) | 60.0 | 51.0 |
| - | Autoencoder+SVM | 67.5(LOSO CV) | 65.0 | 70.0 |
| - | Autoencoder+SVM(TL) | 72.5(LOSO CV) | 75.0 | 70.0 |
| Proposed algorithm | PD_CSTLOK&S2 | **95.0**(LOSO CV) | 90.0 | **100.0** |

To fully illustrate the advantages of the proposed algorithm, the transfer learning(denoted as “algorithm(TL)”) and nontransfer-learning versions of DBN and the autoencoder are also compared, denoted as DBN+SVM, DBN+SVM(TL), Autoencoder+SVM, and Autoencoder+SVM(TL), respectively. For the transfer learning version, the autoencoder and DBN are used for unsupervised learning of the source dataset, and the networks are trained to represent the features of the target dataset; then, one hidden layer of networks is extracted to construct feature sets. Finally, an SVM is used for classification. The source domain dataset is not used for the nontransfer learning version.

The main reasons for choosing to compare the algorithms listed above are as follows: 1) Deep learning is a popular method for feature learning and classification; thus, the representative DBN, CNN and autoencoder methods are chosen. 2) The transfer learning versions of the DBN and autoencoder take advantage of the source domain dataset, and they are more compelling than the compared algorithms.

As shown in Table II, the average accuracy rate of PD_CSTLOK&S2 reached 95% and achieved better results than other state-of-the-art methods.

Table III shows the classification and comparison results of this algorithm are presented. The dataset used is the MaxLittle dataset. The algorithm proposed in this paper is compared with the other representative algorithms on this dataset. In addition, the proposed algorithm is compared with the most relevant algorithms, including the SVM with linear and radial kernels, DBN, CNN and the deep autoencoder algorithm.

TABLE III
COMPARISON OF THE CLASSIFICATION RESULTS OF THE PROPOSED ALGORITHM(MAXLITTLE DATASET)

| Study | Method | Accuracy (%) | Sensitivity(%) | Specificity (%) |
|---|---|---|---|---|
| Little *et al.*[8] | Preselection filter+exhaustivesearch+SVM | 91.40(Bootstrap with 50 replicates) | — | — |
| Shahbaba and Neal [59] | Dirichlet process mixtures | 87.7(5-fold CV) | — | — |
| Psorakis *et al.*[60] | mRVMs | 89.47(10-fold CV) | — | — |
| Guo *et al.*[30] | GA-EM | 93.10(10-fold CV) | — | — |
| Sakar and Kursun [28] | Mutual information+SVM | 92.75(Bootstrap with 50 replicates) | — | — |
| Das [26] | ANN decision tree | 92.90(Holdout) | — | — |
| Ozcift and Gulten[61] | Correlation-based feature selection-rotation forest | 87.10(10-fold CV) | — | — |
| Luukka [45] | Fuzzy entropy measures+similarity | 85.03(Holdout) | — | — |
| Li *et al.*[46] | Fuzzy-based nonlinear transformation+SVM | 93.47(Holdout) | — | — |
| Spadoto *et al.*[31] | PSO+OPF harmony search+OPF gravitational search+OPF | 84.01(Holdout) | — | — |
| Polat [34] | FCMFW+KNN | 97.93(Holdout) | — | — |
| Chen *et al.*[21] | PCA-fuzzy KNN | 96.07(10-fold CV) | — | — |
| Ali *et al.*[42] | DBN | 94(Holdout) | — | — |
| Åström and Koker [27] | Parallel ANN | 91.20(Holdout) | 90.5 | 93.0 |
| Daliri[62] | SVM with Chi-square distance | 91.20(Holdout) | 91.71 | 89.92 |
| | kernel | | | |
| Zuo *et al.*[32] | PSO-fuzzy KNN | 97.47(10-fold CV) | 98.16 | 96.57 |
| Kadam and Jadhav [43] | FESA-DNN | 93.84(10-fold CV) | 95.23 | 90.00 |
| Ma *et al.*[63] | SVM-RFE | 96.29(10-fold CV) | 95.00 | 97.50 |
| Cai *et al.* [37] | RF-BFO-SVM | 97.42(10-fold CV) | 99.29 | 91.5 |
| Dash *et al.*[64] | ECFA-SVM | 97.95(10-fold CV) | 97.90 | — |
| Gürüler [44] | KMCFW-CVANN | 99.52(10-fold CV) | 100 | 99.47 |
| - | SVM(linear kernel) | 75.0(LOSO) | 100.0 | 0.0 |
| - | SVM(RBF kernel) | 75.0(LOSO) | 100.0 | 0.0 |
| Proposed algorithm | PD_CSTLOK&S2 | **100.0**(LOSO) | **100.0** | **100.0** |

As shown in Table III, the compared methods on the Max Little dataset are based on hold-one-out CV and 10-fold CV, Holdout CV is more contingent, and even when 10-fold CV is adopted, there is still no deliberate effort to avoid the fact that the training samples and test sample come from the same subject. Therefore, the accuracies of the methods are perhaps higher than they would be in practice. As TABLE III shows, even under LOSO CV, the proposed algorithm still achieves the best performance.

It can also be found from Table IV that the proposed algorithm achieves the best results on the DNSH dataset. Outperforming the SVM, the average classification accuracy of the proposed algorithm reaches 86.7%, proving that it is quite effective even on the DNSH dataset.

TABLE IV
COMPARISON OF THE CLASSIFICATION RESULTS OF THE PROPOSED ALGORITHM(DNSH DATASET)

| Study | Method | Accuracy (%) | Sensitivity(%) | Specificity (%) |
|---|---|---|---|---|
| - | SVM(linear kernel) | 61.3 | 0.0 | 90.5 |
| - | SVM(RBF kernel) | 67.7 | 0.0 | 100.0 |
| Proposed algorithm | PD_CSTLOK&S2 | 86.7 | 100.0 | 80.0 |

### *E. Time Complexity Analysis*

The time cost of the proposed algorithm is another important aspect of its practical feasibility. Fig. 4 shows the time cost of feature extraction from one subject of the Max Little dataset. The abscissa is the number of iterations, and the ordinate is the time cost. As the number of iterations increases, the time cost of the proposed method increases less rapidly than does that of the fast convolutional sparse coding method in [65]. This result means that the engineering complexity of the proposed algorithm is acceptable.

Table V shows the time cost of the proposed algorithm based on the number of convolution kernels. The kernel training time was ignored because the optimal kernel is fixed.

As Table V shows, the time cost increases only slowly as the number of convolution kernels increases. On the MaxLittle and DNSH datasets, the time cost increases slowly with the number of kernels; on the Sakar dataset, the subject data size ($H_0 \times N$)is larger; consequently,the number of kernels is not the key factor of the computational complexity; thus, the time cost is not strictly increased: the minimum time cost is 2.93 s, the maximum time cost is 3.83 s, and the difference is 0.09 s, indicating that the time cost is acceptable even under large numbers of kernels.

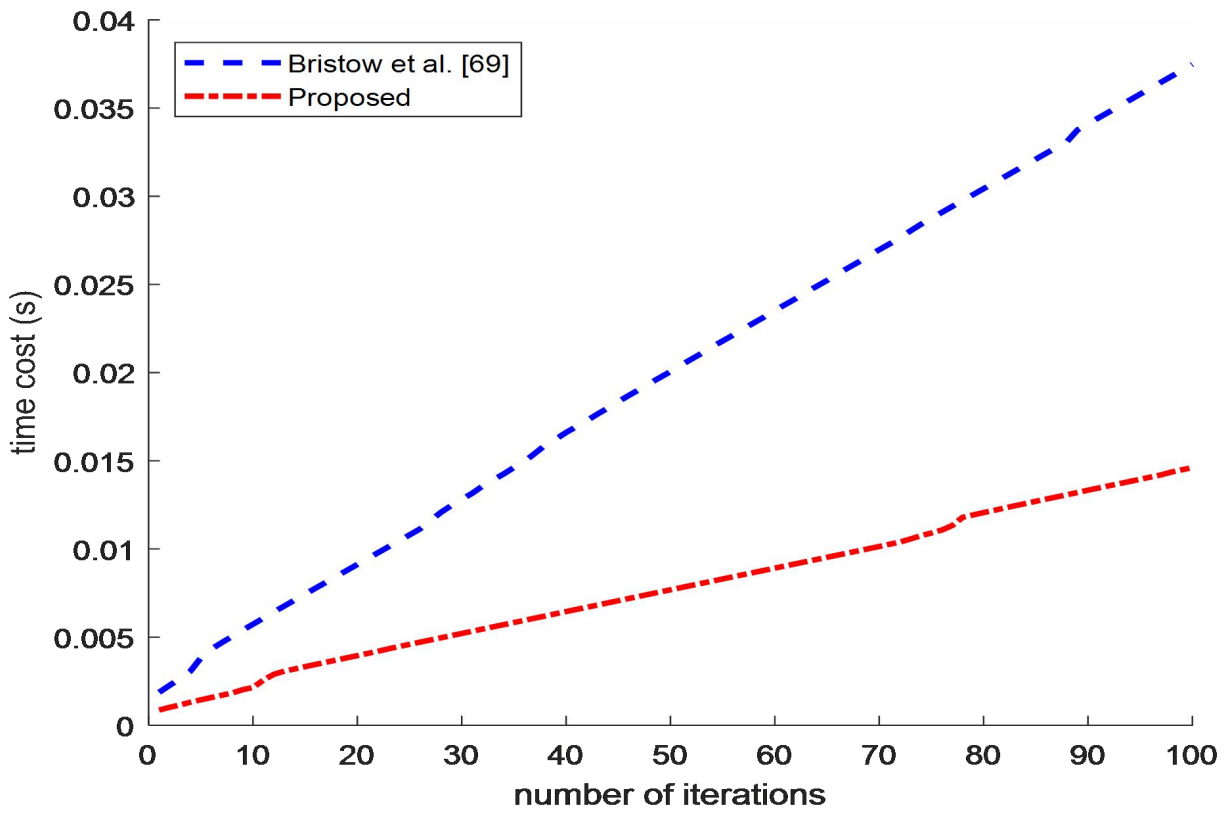

Fig. 4.Time cost of feature extraction from one subject of the Max Little dataset.

TABLE V
THE TRAINING TIME COST OF THE PROPOSED ALGORITHM BASED ON THE NUMBER OF CONVOLUTION KERNELS

| Sackar dataset | | MaxLittle dataset | | DNSH dataset | |
|---|---|---|---|---|---|
| convolution kernel num | Time cost(s) | convolution kernel num | Time cost(s) | convolution kernel num | Time cost(s) |
| 2 | 27.46 | 2 | 1.31 | 2 | 2.93 |
| 3 | 28.43 | 3 | 1.33 | 3 | 3.05 |
| 4 | 28.94 | 4 | 1.34 | 4 | 3.24 |
| 5 | 28.83 | 5 | 1.37 | 5 | 3.27 |
| 6 | 29.06 | 6 | 1.37 | 6 | 3.40 |
| 7 | 29.03 | 7 | 1.43 | 7 | 3.57 |
| 8 | 28.48 | 8 | 1.44 | 8 | 3.83 |

## IV. DISCUSSION AND CONCLUSIONS

The current relevant PD diagnostic methods of speech data are primarily based on local PD data; however, the problem of insufficient samples prevents the methods from improving their classification accuracy. To solve this problem, a new method is proposed in this paper: a convolution sparse learning algorithm is adopted to reduce the dimensions and the demand for larger numbers of samples based on sparse transfer learning from public datasets and kernel optimization; mixed speech feature learning is used to optimize the sample features and improve the classification accuracy. By the end of this study, tens of representative algorithms were applied to verify the performance and for comparison with the proposed algorithm. The experimental results show that the innovative parts of the proposed algorithm, including convolution sparse transfer learning, kernel optimization, and parallel selection, are effective. Compared with other state-of-the-art algorithms, the proposed algorithm in this paper achieves significant improvements in terms of classification accuracy, sensitivity and specificity. In addition, the time complexity of the proposed algorithm is acceptable.

The main contributions and innovations of this paper are as follows: 1) Convolution sparse coding and transfer learning are combined to solve the small sample problem that exists with currently available PD speech data. 2)A parallel sample/feature selection mechanism is designed to follow the convolution sparse coding and transfer learning operations to further improve the quality of the samples and features. 3) A convolution kernel optimization mechanism is proposed to enhance the current CSC algorithm. 5) Three representative PD speech datasets are considered to study and verify the classification algorithm.

Because the available PD training datasets have small sample sizes, a major contradiction always exists. Based on the research results of this paper, the next steps are to conduct further studies regarding the size and type of public speech datasets and to investigate various domain adaptation architectures.

## ACKNOWLEDGMENT

We are grateful for the support of the National Natural Science Foundation of China NSFC (No. 61771080); the Fundamental Research Funds for the Central Universities (2019CDCGTX306, 2019CDQYTX019); Basic and Advanced Research Project in Chongqing(cstc2018jcyjAX0779, cstc2018jcyjA3022);Chongqing Social Science Planning Project(2018YBYY133).

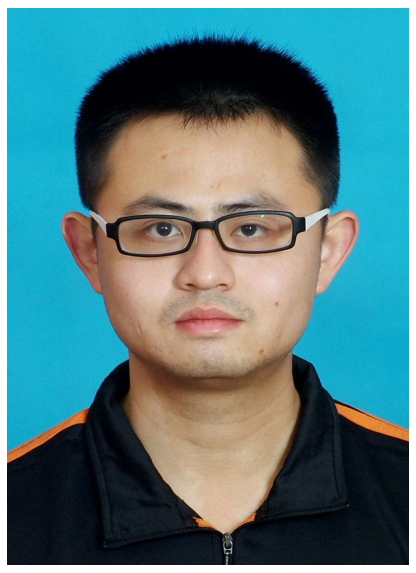

**XiaohengZhang** received an M.S. degree in Circuits and Systems from Chongqing University, China, in 2007. He is an engineer with many years of engineering experience in signal processing.
He is currently an associate professor with Chongqing Radio and TV University. His current research interests include speech signal processing, pattern recognition and machine learning.

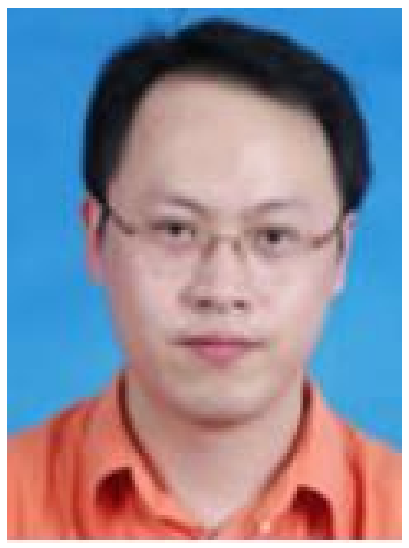

**Yongming Li** received M.S. and Ph.D. degrees in Circuits and Systems from Chongqing University, China, in 2003 and 2007, respectively. He was a visiting scholar at Pennsylvania State University, USA from 2008 to 2009, and worked as a postdoctoral fellow at Carnegie-Mellon University, USA, from Feb. to Dec. 2009. He is currently a professor with the School of Microelectronics and Communication Engineering, Chongqing University and serves as an editorial board member for IEEE Access. His current research interests include signal processing, pattern recognition, data mining, and machine learning.

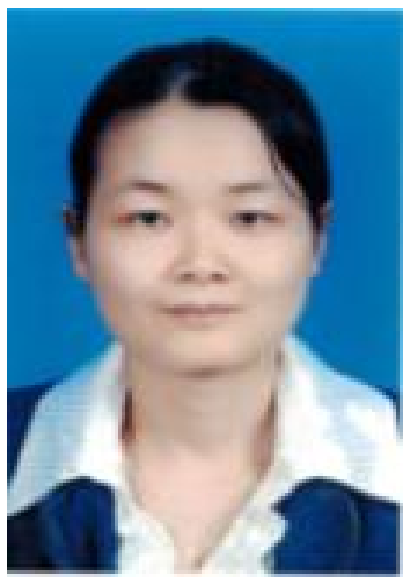

**Pin Wang** received an M.S. degree from Chongqing University, China, in 2003 and a Ph.D. degree from Nanyang Technological University, Singapore, in 2008. She worked as a postdoctoral at Nanyang Technological University, Singapore, from 2008 to 2009. In 2009–2010, she was funded by the National Energy Laboratory Postdoctoral Fund to conduct postdoctoral research at the University of Pittsburgh.
She is currently an associate professor at Chongqing University. Her current research interests include hyperspectral imaging and detection, intelligent information processing, and big data analysis.

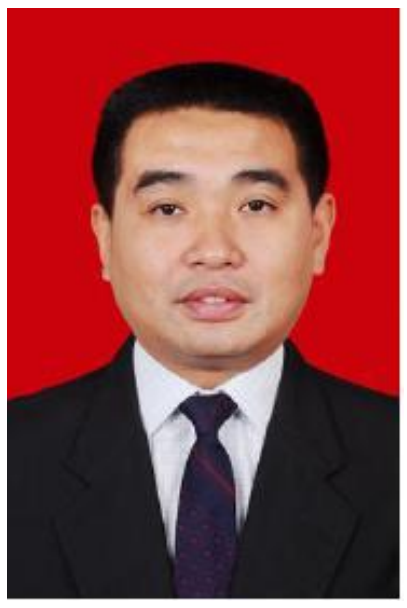

**Xiaoheng Tan** received the B.E. and Ph.D. degrees in electrical engineering from Chongqing University, Chongqing, China, in 1998 and 2003, respectively. He was a Visiting Scholar with the University of Queensland, Brisbane, Qld., Australia, from 2008 to 2009. He is currently a professor with the School of Microelectronics and Communication Engineering, Chongqing University. His current research interests include modern communications technologies and systems, communications signal processing, pattern recognition, and machine learning.

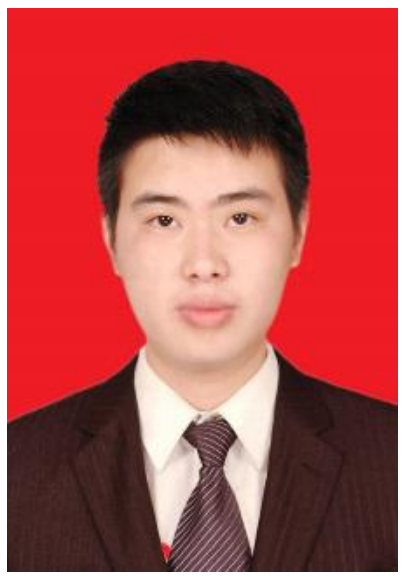

**Yuchuan Liu** received a bachelor's degree incommunication engineering from the Southwest University of Science and Technology, China, in 2015.He is currently pursuing a Ph.D. degree with Chongqing University, Chongqing, China. His research interests include dimensionality reduction, pattern recognition, and machine learning.